\documentclass[acmtog]{acmart}
\AtBeginDocument{%
  }

\acmJournal{TOG}

\acmSubmissionID{813}

\usepackage{pifont}
\newcommand{\cmark}{\ding{51}}
\newcommand{\xmark}{\ding{55}}
\newcommand{\md}[1]{\textcolor[rgb]{0.0,0.0,0.0}{#1}}

\copyrightyear{2026}
\acmYear{2026}
\setcopyright{cc}
\setcctype{by}
\acmConference[SIGGRAPH Conference Papers '26]{Special Interest Group on Computer Graphics and Interactive Techniques Conference Conference Papers}{July 19--23, 2026}{Los Angeles, CA, USA}
\acmBooktitle{Special Interest Group on Computer Graphics and Interactive Techniques Conference Conference Papers (SIGGRAPH Conference Papers '26), July 19--23, 2026, Los Angeles, CA, USA}
\acmDOI{10.1145/3799902.3811138}
\acmISBN{979-8-4007-2554-8/2026/07}
\begin{document}

\title{VidSplat: Gaussian Splatting Reconstruction with Geometry-Guided Video Diffusion Priors}

\author{Jimin Tang}
\authornote{Both authors contributed equally to this research.}
\affiliation{%
  \institution{School of Software, Tsinghua University}
  \city{Beijing}
  \country{China}
}
\email{tangjm24@mails.tsinghua.edu.cn}

\author{Wenyuan Zhang}
\authornotemark[1]
\affiliation{%
  \institution{School of Software, Tsinghua University}
  \city{Beijing}
  \country{China}
}
\email{zhangwen21@mails.tsinghua.edu.cn}

\author{Junsheng Zhou}
\affiliation{%
  \institution{School of Software, Tsinghua University}  
  \city{Beijing}
  \country{China}
}
\email{zhou-js24@mails.tsinghua.edu.cn}

\author{Zian Huang}
\affiliation{%
  \institution{School of Software, Tsinghua University}  
  \city{Beijing}
  \country{China}
}
\email{huangza25@mails.tsinghua.edu.cn}

\author{Kanle Shi}
\affiliation{%
  \institution{Kuaishou Technology}
  \city{Beijing}
  \country{China}
}
\email{shikanle@kuaishou.com}

\author{Shenkun Xu}
\affiliation{%
  \institution{Kuaishou Technology}
  \city{Beijing}
  \country{China}
}
\email{xushenkun@kuaishou.com}

\author{Yu-Shen Liu}
\authornote{The corresponding author is Yu-Shen Liu.}
\affiliation{%
  \institution{School of Software, Tsinghua University}  
  \city{Beijing}
  \country{China}
}
\email{liuyushen@tsinghua.edu.cn}

\author{Zhizhong Han}
\affiliation{%
  \institution{Department of Computer Science, Wayne State University}
  \city{Detroit}
  \country{USA}
}
\email{h312h@wayne.edu}

\renewcommand{\shortauthors}{Jimin Tang et al.}

\begin{abstract}
  Gaussian Splatting has achieved remarkable progress in multi-view surface reconstruction, yet it exhibits notable degradation when only few views are available. Although recent efforts alleviate this issue by enhancing multi-view consistency to produce plausible surfaces, they struggle to infer unseen, occluded, or weakly constrained regions beyond the input coverage. To address this limitation, we present \textbf{\md{VidSplat}}, a \md{training-free} generative reconstruction framework that leverages powerful video diffusion priors to \md{iteratively} synthesize novel views that compensate for missing input coverage, and thereby recover complete 3D scenes from sparse inputs. Specifically, we tackle two key challenges that enable the effective integration of generation and reconstruction. First, for 3D consistent generation, we elaborate a training-free, stage-wise denoising strategy that adaptively guides the denoising direction toward the underlying geometry using the rendered RGB and mask images. Second, to enhance the reconstruction, we develop an iterative mechanism that samples camera trajectories, explores unobserved regions, synthesizes novel views, and supplements training through confidence weighted refinement. \md{VidSplat} performs robustly to sparse input and even a single image. Extensive experiments on widely used benchmarks demonstrate our superior performance in sparse-view scene reconstruction. 
  Project Page: \url{https://tangjm24.github.io/VidSplat}.
\end{abstract}


\begin{teaserfigure}
  \centering
  \includegraphics[width=\textwidth]{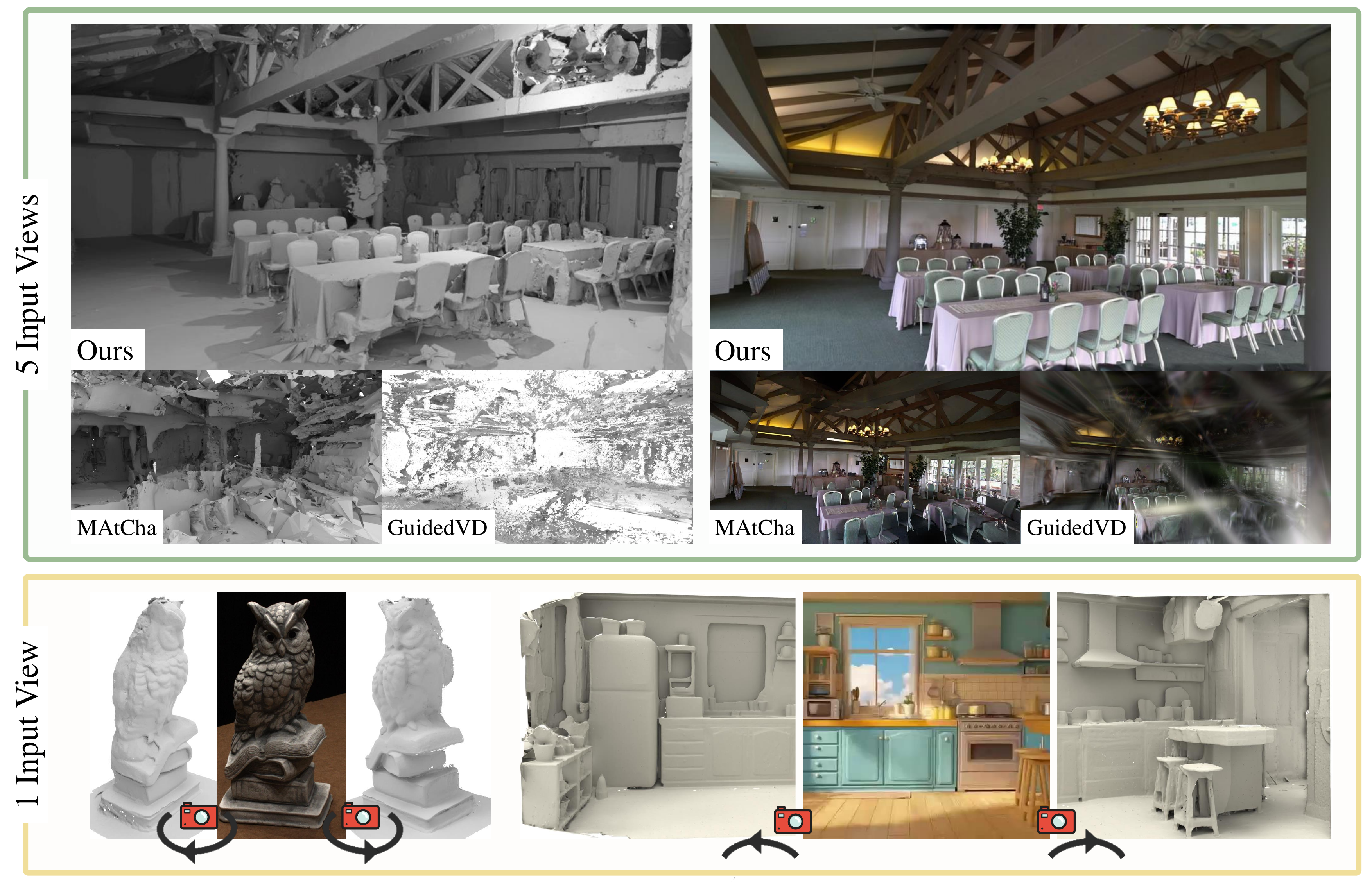}
  \vspace{-0.6cm}
  \caption{We highlight the strength of \textbf{\md{VidSplat}} in large-scale scene reconstruction and novel view synthesis using only 5 input views (top), where recent sparse-view reconstruction methods fail to recover reasonable surfaces. We also demonstrate our ability to generate a complete scene from a single input image, either with all-around coverage (bottom-left) or outward-expanding completion (bottom-right).}
  \label{fig:teaser}
  \Description{We highlight the strength of \textbf{\md{VidSplat}} in large-scale scene reconstruction and novel view synthesis using only 5 input views (top), where recent sparse-view reconstruction methods fail to recover reasonable surfaces.}
\end{teaserfigure}

\maketitle

\section{Introduction}
Reconstructing 3D geometry from multi-view images is a fundamental task in computer vision~\cite{mildenhall2020nerf,kerbl20233dgs, chen2024pgsr,yu2024hifi,fang2026more,zhang2025materialrefgs}, as it lifts 2D observations into 3D representations that enable interaction and manipulation~\md{\cite{yu2024hifi}}, and underpins a wide range of downstream applications such as digital content creation, VR/AR, and embodied intelligence. Recent advances have achieved remarkable progress by learning neural radiance fields (NeRF)~\cite{mildenhall2020nerf} as implicit scene representations or adopting 3D Gaussian Splatting (3DGS)~\cite{kerbl20233dgs} as explicit ones, leading to breakthroughs in surface reconstruction. However, both paradigms degrade notably when only a few input views are available, because their optimization relies heavily on multi-view consistency, which becomes ill-posed and under-constrained in sparse-view settings.


To address this limitation, recent generalizable approaches~\cite{na2024uforecon,younes2024sparsecraft,chang2025meshsplat,liang2024retr} pretrain volumetric representations on large-scale datasets to learn cross-view correspondences, and then infer the unseen scenes for reconstruction. Other scene-specific methods~\cite{wu2023svolsdf,huang2024neusurf} overfit a single scene by introducing various monocular~\cite{han2025sparserecon,guedon2025matcha} or multi-view-stereo~\cite{wu2025sparis,huang2025fatesgs} priors. However, these methods suffer from generalization or scalability to large and complex environments. More critically, they remain constrained to recovering only the visible regions of the given views and cannot infer geometry outside the field of view, which limits their applicability to broader 3D scenarios.


To tackle these issues, we introduce \md{VidSplat}, a  generative reconstruction framework  for recovering complete and high-fidelity 3D scenes from sparse input. Our approach draws inspiration from recent advances in general video diffusion models~\cite{wan2025wan,gao2025seedance,zhang2025waver}, which are pretrained on large-scale video datasets and thus inherently encode rich geometric priors over diverse scene appearances and viewpoints. Specifically, we generate video clips conditioned on sampled camera trajectories and reference images~\cite{yu2024viewcrafter,hou2025camtrol} to expand the sparse view coverage of the scene. To promote 3D consistency across the synthesized sequences, we propose a training-free, stage-wise denoising strategy that leverages rendered RGB and mask images at each view to guide the denoising toward the underlying geometry. At higher noise levels, the denoising is constrained to follow RGB signals within masked regions which suppresses dynamics and content drift. At lower noise levels, this constraint is gradually relaxed, enabling the model to refine imperfect renderings for coherent 3D reconstruction.


We further introduce several techniques to seamlessly integrate the generated results into the reconstruction pipeline. We first develop a visibility-based camera pose sampling strategy that navigates from the existing views toward insufficiently covered regions, which are identified to require additional view synthesis. We then introduce trajectory expansion and view selection strategies to mitigate hallucinations of the video model. Finally, the synthesized results are incorporated into the training process through confidence-weighted fusion, and the reconstruction is iteratively refined to progressively recover complete 3D scenes with smooth and high-fidelity geometric details.


We extensively evaluate \md{VidSplat} on diverse real-world datasets covering both indoor and outdoor scenarios, where we achieve state-of-the-art performance in both surface reconstruction and novel view synthesis. We also demonstrate strong generative capability of our framework from a single input view, as highlighted in Fig.~\ref{fig:teaser}. In summary, our main contributions are as follows:

\begin{itemize}
    \item We propose a generative surface reconstruction framework from sparse input views with video diffusion priors, which iteratively incorporates video generation into reconstruction for continuous refinement.
    \item We introduce a training-free, stage-wise denoising strategy that adaptively guides the denoising direction toward underlying geometry for 3D consistent video generation. 
    \item We achieve state-of-the-art results in widely adopted real-world benchmarks for both surface reconstruction and novel view synthesis.
\end{itemize}

\section{Related Work}
\subsection{Sparse-view Surface Reconstruction}
Recently, NeRF~\cite{mildenhall2020nerf} and 3DGS~\cite{kerbl20233dgs} have become paradigms for 3D surface reconstruction~\cite{huang20242dgs,zhang2024gspull,chen2024pgsr,li2025vags,li2025gaussianudf,zhang2026vrpudf,noda2026gsprior,zhou2026udfstudio}. However, their optimization relies on photometric consistency across dense views and degrades significantly under sparse inputs. Recent solutions can be categorized into two directions. Generalizable methods pretrain networks on large-scale datasets to capture cross-view patterns and then generalize to unseen scenes~\cite{na2024uforecon,younes2024sparsecraft,chang2025meshsplat}. 
Overfitting methods optimize the specific scene from sparse inputs~\cite{han2025sparserecon,huang2025fatesgs,guedon2025matcha,wu2025sparse2dgs} by incorporating geometric priors such as point clouds~\cite{han2025sparserecon,wu2025sparse2dgs,li2025ifiltering,chen2024neuraltps}, normals~\cite{zhang2025monoinstance,ni2025g4splat,li2025pffnet}, local patterns~\cite{raj2024spurfies}, or exploiting multi-view cues like semantic features~\cite{huang2025fatesgs,wu2023svolsdf} or manifolds~\cite{guedon2025matcha}. Despite these efforts, their reconstructions remain limited to the visible regions of the input views and cannot infer geometry beyond them, which results in incomplete and fragmented surfaces under sparse view conditions.

\begin{figure*}[t]
    \centering
    \includegraphics[width=\linewidth]{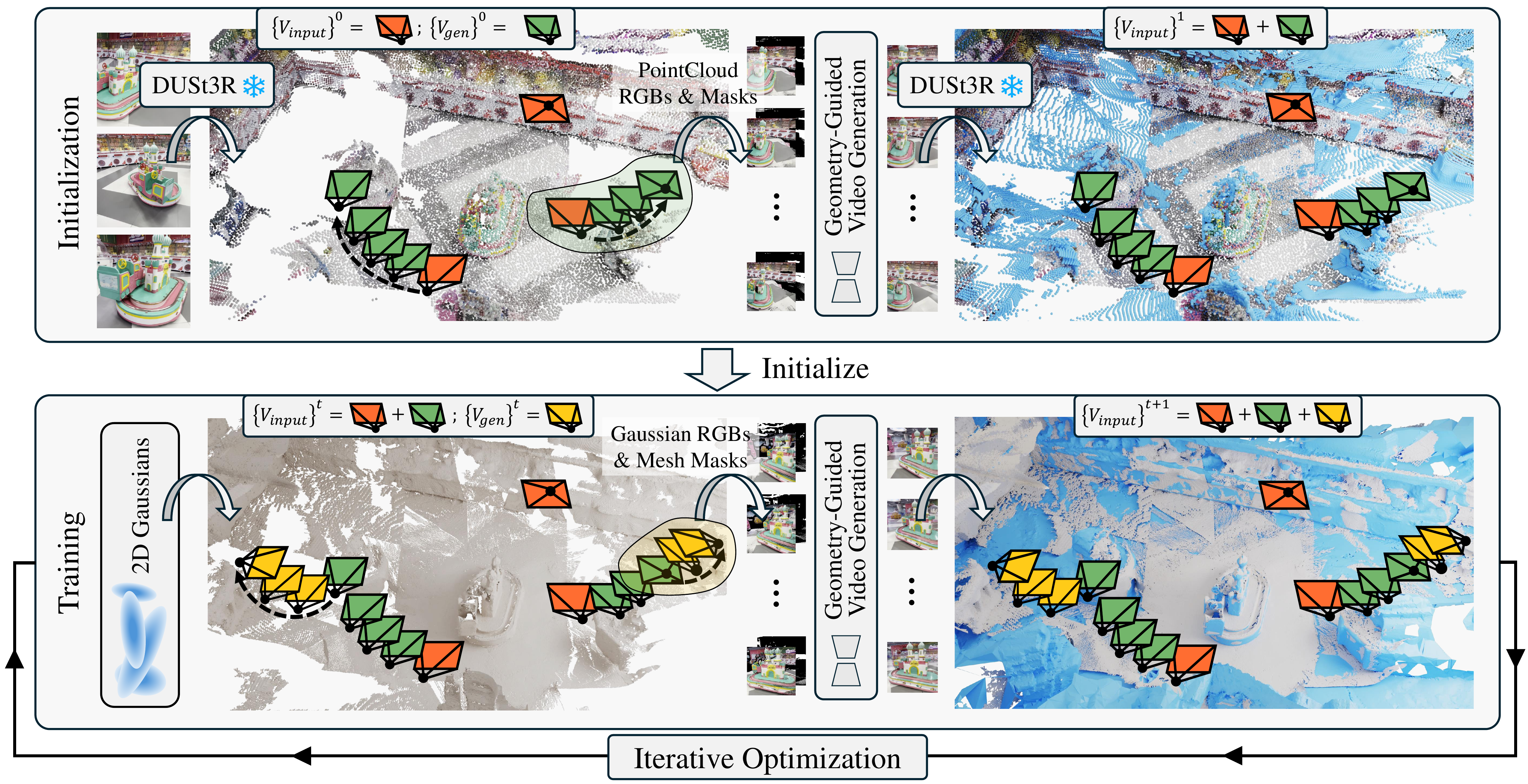}
    \vspace{-0.5cm}
    \caption{Overview of our optimization framework. Given sparse input views, we sample novel camera trajectories and employ a camera-controlled video diffusion model (VDM) with our geometry-guided denoising strategy to generate additional views. In the initialization stage, RGB and mask images rendered from point cloud are used as VDM inputs, and the generated views are used to complete the initial point cloud. In the training stage, Gaussian-rendered RGBs and mesh-rendered masks are used as inputs, and the generated views are used to expand the training view set. The newly added point clouds and mesh surfaces are highlighted in blue.}
    \label{fig:overview}
    \Description{Overview of our optimization framework.}
    \vspace{-0.2cm}
\end{figure*}

\subsection{Novel View Synthesis from Sparse Inputs}
3DGS has demonstrated remarkable advantages in quality and efficiency for novel view synthesis~\cite{kerbl20233dgs,lu2024scaffold,zhang2026gaussiangrow}. However, similar to the challenges in surface reconstruction, its performance depends on the number of input views~\cite{han2024binocular,zhou20264c4d,huang2025fatesgs,xiang2026vggs}.
More recent studies introduce generative priors for additional supervisions from novel views. Although these methods are conceptually related to our work, they have three key limitations. First, they require pretraining large-scale image-to-image (I2I)~\cite{paliwal2025ri3d,wu2025difix3d+,kong2025gsgs,fischer2025flowr,wei2025gsfix3d} or video-to-video (V2V)~\cite{wu2025genfusion,yin2025gsfixer,ma2025see3d,liu20243dgsenhancer} diffusion models, which entails substantial computational cost. Second, while they can repair artifacts at interpolated viewpoints, they fail to recover unseen regions at extrapolated views, where the rendering often appear as voids. Third, being tailored for novel view synthesis~\cite{zhong2025taming,wu2025difix3d+}, they produce visually plausible renderings but lack consistent underlying geometries. Our method belongs to this category, and addresses these limitations, differing the previous methods a lot.


\subsection{Controllable Video Diffusion Models}
Breakthroughs in image diffusion models~\cite{ho2020ddpm,dhariwal2021diffusion} have fueled rapid progress in video generation. Scalable training paradigms based on conditional denoising have enabled controllable video synthesis, with applications such as audio-driven avatar animation~\cite{ding2025klingavatar,gao2025wans2v} and direction-conditioned world modeling~\cite{yu2025gamefactory}. Of particular relevance to our work is camera-controlled video generation~\cite{wang2024motionctrl,yu2024viewcrafter,hou2025camtrol,he2025cameractrlii}. To be specified, CameraCtrl~\cite{he2025cameractrl} encodes camera motion into the attention layers of a U-Net backbone. TrajectoryCrafter~\cite{mark2025trajectorycrafter} warps the input view along predefined camera paths for reference video conditioning, while CamTrol~\cite{hou2025camtrol} employs the inversion of point cloud renderings to offer layout priors for generation. Although effective for open-domain content creation, these approaches often produce dynamics and shakes that harm 3D geometry consistency. 
\md{This limitation motivates us to develop a geometry-aware video diffusion framework with explicit camera control, where denoising is guided by rendered geometry and the generated results are further incorporated into iterative reconstruction for consistent and complete 3D scene recovery.}

\section{Method}

Given a set of sparse input views of a scene, we aim to reconstruct complete and high-quality scene surfaces. We start by initializing 3D points from input views and newly sampled views using DUSt3R~\cite{wang2024dust3r}.
With the 3D points, we initialize 2D Gaussians, then train 2D Gaussians~\cite{huang20242dgs,guedon2025matcha} by repeating the aforementioned procedures iteratively, where we use 2DGS to render novel views and use the video diffusion model to inpaint the unseen regions on novel views. An overview of our method is illustrated in Fig.~\ref{fig:overview}.


\subsection{Preliminary}
\label{sec3.1}
3D Gaussian Splatting (3DGS)~\cite{kerbl20233dgs} has become a paradigm for learning 3D representations from multi-view images. A scene is modeled as a set of learnable anisotropic Gaussian primitives $\{G_i\}_{i=1}^K$, each with attributes like position $x_i$, opacity $o_i$, and color $c_i$. We can obtain RGB images by rasterizing Gaussians in a splatting manner,
\begin{equation}
    C=\sum_{i=1}^K c_i*p_i*o_i*\prod_{j=1}^{i-1}(1-o_i),
\end{equation}
where $p_i$ is the 2D kernel of the projected $G_i$. The Gaussian parameters are then optimized by the supervision of ground truth views. Recent 2DGS~\cite{huang20242dgs} flattens 3D Gaussians into disks, which promotes alignment between Gaussians and surfaces and thus improves the geometry fidelity.

Video Diffusion Models synthesize videos by learning a conditional generative process that maps Gaussian noise to natural sequences. Early systems use U-Net backbones with spatiotemporal convolutions~\cite{blattmann2023align,blattmann2023stable}, whereas recent Diffusion Transformer architectures (DiT) have demonstrated stronger scalability and effectiveness~\cite{wan2025wan,yang2025cogvideox}. In terms of training objectives, rather than DDPM-style reverse process via SDE/ODE solvers, flow matching~\cite{lipman2022flowmatching} has recently become a mainstream alternative. Given a data sample $x_0\sim p_{data}(x)$, a forward interpolation can be written as 
\begin{equation}
    x_t=(1-t)x_0+t\epsilon, \epsilon\sim \mathcal{N}(0, I), t\in[0,1].
\end{equation}
\noindent The model learns a parametric velocity field $v_\theta$ by minimizing the objective
\begin{equation}
    \mathcal{L}(\theta)=\mathbb{E}_{x_0, \epsilon, t}\Vert v_\theta(x_t, t, c) - v^* \Vert^2,
\end{equation}
\noindent where $v^*=\frac{dx(t)}{dt}=\epsilon-x_0$ is the target constant velocity.

\subsection{Optimization Framework}
\label{sec3.2}
We denote $\{V_{input}\}^t$ and $\{V_{gen}\}^t$ as the input and newly generated views at $t$-th cycle, following 
\begin{equation}
    \{V_{input}\}^{t+1}=\{V_{input}\}^t\cup\{ V_{gen}\}^t.
\end{equation}
\noindent The first update $\{V_{input}\}^0\!\rightarrow\!\{V_{input}\}^1$ is performed once during initialization, while subsequent updates $\{V_{input}\}^t\!\rightarrow\!\{V_{input}\}^{t+1}$ are iterated during Gaussian training.

As illustrated in Fig.~\ref{fig:overview}, given sparse input views $\{V_{input}\}^0$, we first construct an initial point cloud using DUSt3R~\cite{wang2024dust3r}. Based on $\{V_{input}\}^0$, we sample visibility-based camera trajectories (Sec.~\ref{sec3.3}) and render the point cloud into RGB and mask images, which are fed into the video diffusion model to synthesize 3D consistent video clips (Sec.~\ref{sec3.4}). A subset of keyframes from the generated sequences forms $\{V_{gen}\}^0$, which is merged with $\{V_{input}\}^0$ to obtain $\{V_{input}\}^1$ and to rerun DUSt3R, yielding a denser point cloud for Gaussian initialization.

During Gaussian training, we perform multiple refinement cycles. In each cycle $t$, we sample new trajectories and generate new sequences via video diffusion model to obtain $\{V_{gen}\}^t$, which are merged with $\{V_{input}\}^t$ into $\{V_{input}\}^{t+1}$ to expand the view coverage. RGB images are rendered via Gaussian rasterization, while masks are computed from ray tracing on periodically evaluated meshes rather than Gaussian-rendered alpha maps~\cite{zhong2025taming,paliwal2025ri3d}, which often produce artifacts in unseen regions due to oversized primitives. Since $\{V_{gen}\}^t$ cannot be used for re-initialization, we instead use them to create additional Gaussian centers by backprojecting them into 3D space via monocular depths, following ~\cite{wu2025genfusion}. \md{The newly added Gaussians may not perfectly align with the existing ones initially due to depth estimation limitation, but their positions are progressively refined and become well-aligned during the optimization.}
After the optimization, we use marching tetrahedra~\cite{yu2024gof} to extract the final surfaces.

\begin{figure}[t]
    \centering
    \includegraphics[width=\linewidth]{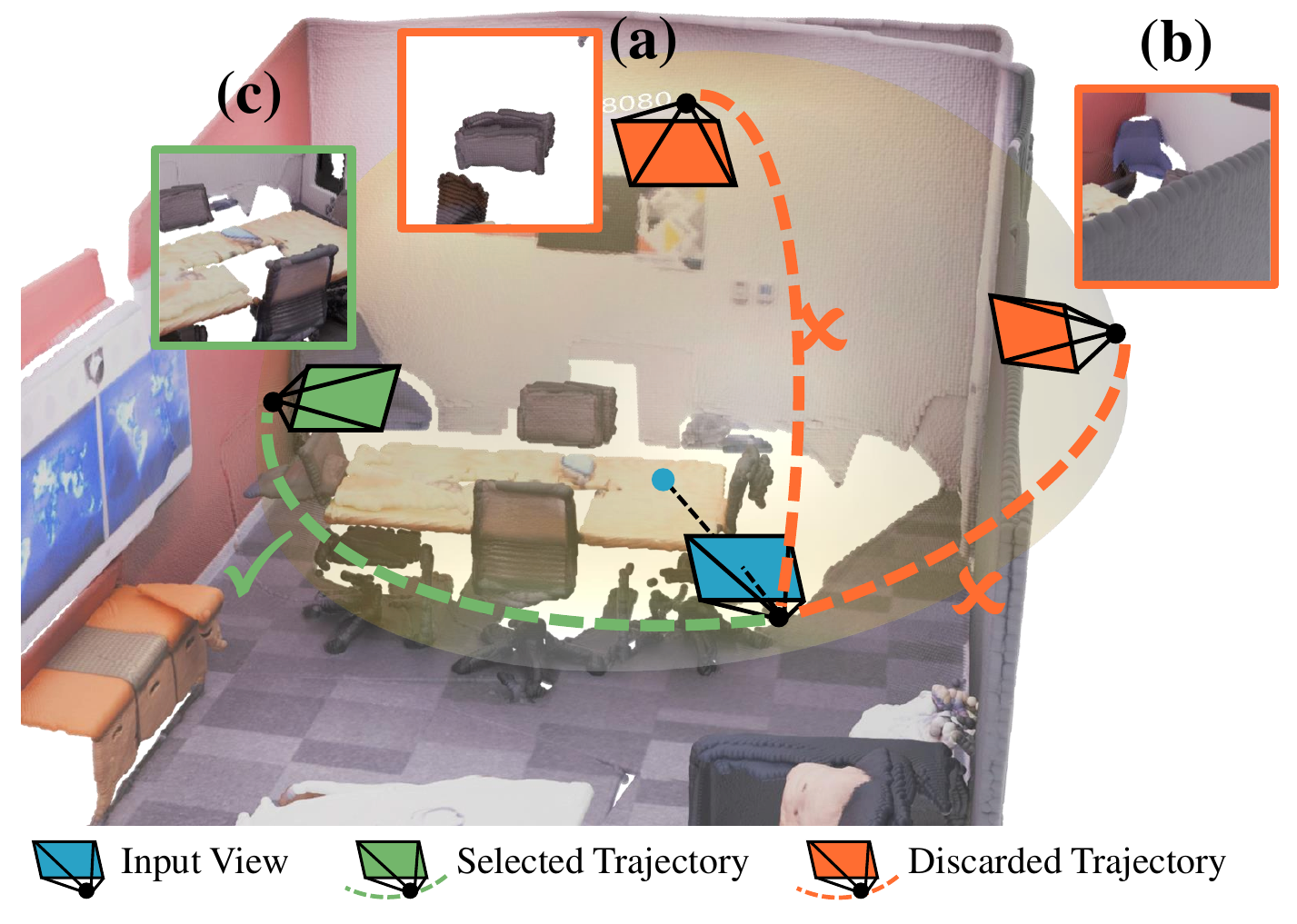}
    \vspace{-0.7cm}
    \caption{Illustration of our visibility-based camera pose sampling strategy. Camera trajectories are constructed on a spherical surface, and the visibility of keyframes is evaluated accordingly. For example, (a) is discarded due to excessive unseen region coverage, while (b) is discarded because of occlusion by the wall.}
    \vspace{-0.5cm}
    \label{fig:method-trajectory}
    \Description{Illustration of our visibility-based camera pose sampling strategy.}
\end{figure}

\subsection{Visibility-Based Camera Pose Sampling}
\label{sec3.3}
Selecting appropriate camera trajectories is critical for exploring under-covered regions. The trajectories should capture as much novel information as possible while preserving reliable geometric references. Existing methods~\cite{zhong2025taming,wu2024reconfusion,yin2025gsfixer} typically construct interpolated or circular paths from input views, which cannot adapt to diverse scene layouts or effectively explore unseen regions. To overcome this limitation, we propose a novel \textit{visibility-based camera pose sampling strategy} to prompt more views to cover larger areas, as illustrated in Fig.~\ref{fig:method-trajectory}. For an input view, we find the intersection point between the camera ray and the scene surface, and construct multiple trajectories where the camera orbits this point on a sphere. \md{Here, a trajectory refers to a virtual camera path constructed beyond the input views, along which the video diffusion model generates what the camera should observe.} The eligibility of each trajectory is evaluated using the the depths $D_i$ and masks $M_i$ of its keyframes. A trajectory is valid only if its views are free from near-plane occlusions and have an appropriate coverage of unseen regions:
\begin{equation}
\label{eq:criteria}
    S_{low} < \text{area}(M_i) < S_{high}, \ \min_{p\in\Omega}D_i(p)>d_0,
\end{equation}
\noindent where $S_{low}, S_{high}, d_0$ are predefined hyperparameters, respectively. \md{The near-plane occlusion refers to the case where the camera moves beyond the scene boundary, causing the view to be blocked by walls or grounds, rather than capturing the scene from a close distance.} For instance, we sample three candidate trajectories in Fig.~\ref{fig:method-trajectory} (a), (b), (c). Per Eq.~\ref{eq:criteria}, we will use the trajectory (a), and discard the other two in (b) and (c). 


We then render RGB and mask images along the selected trajectories for camera-controlled video generation. We deliberately extend the trajectory by 25\% before generation and discard these additional frames afterward, because we observe that the tail frames often exhibit hallucinations caused by error accumulation. From the remaining sequences, we select keyframes that are visually sharp and have large pose variations. These novel views are subsequently used for point cloud estimation during initialization, and as additional supervision during training, respectively.

\begin{figure}
    \centering
    \includegraphics[width=\linewidth]{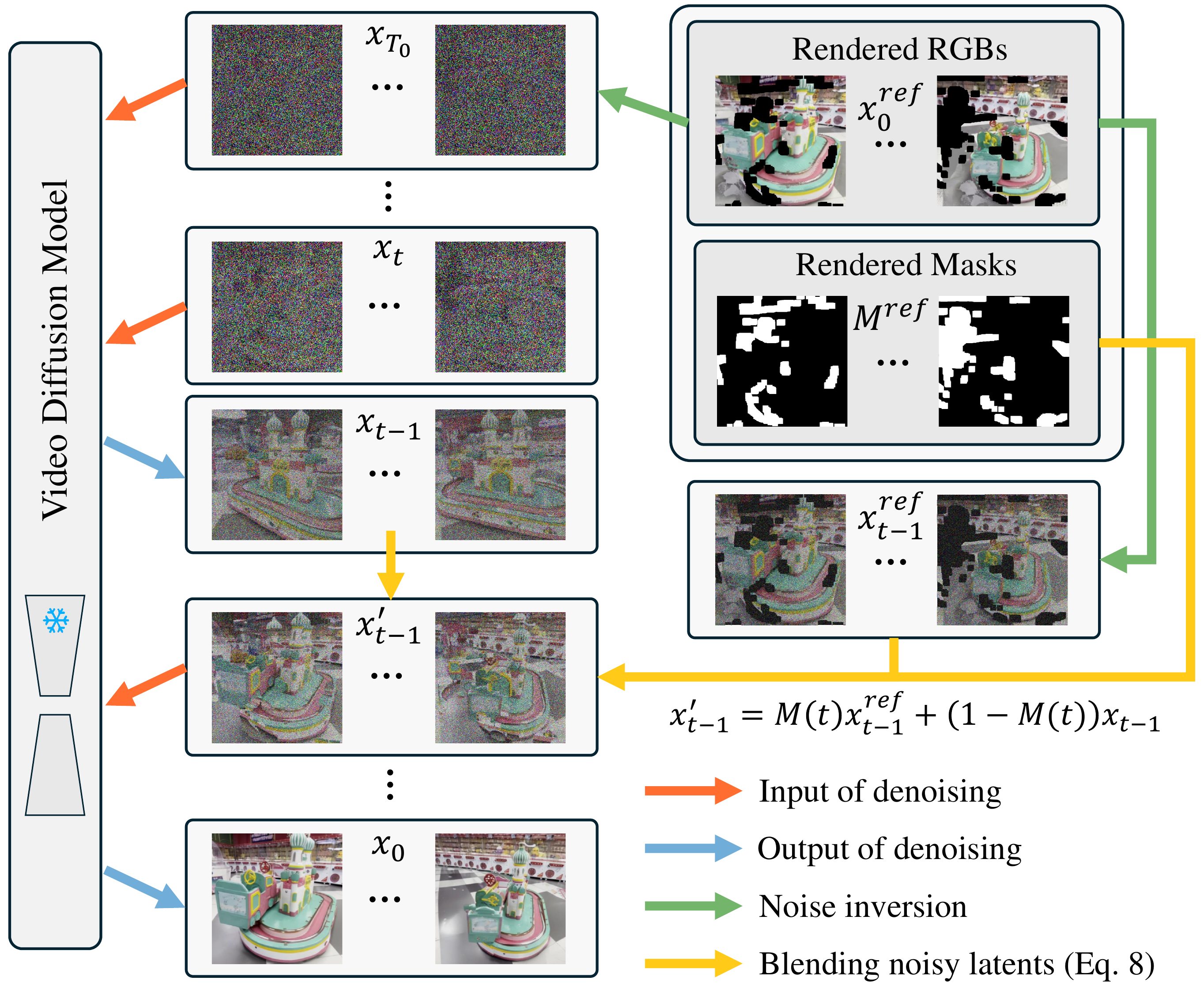}
    \vspace{-0.7cm}
    \caption{Illustration of our geometry-guided denoising for 3D consistent video generation. At each step, we blend the noisy latents $x_{t-1}$ with the reference inversion $x_{t-1}^{ref}$ using $M(t)$ to guide the denoising direction toward underlying geometry.}
    \vspace{-0.4cm}
    \label{fig:denoising}
    \Description{Illustration of our geometry-guided denoising for 3D consistent video generation.}
\end{figure}

\subsection{Geometry-Guided Video Generation}
\label{sec3.4}
We adopt a training-free camera-controlled generation strategy~\cite{hou2025camtrol}, as illustrated in Fig.~\ref{fig:denoising}. Specifically, we construct a sequence of noisy latents by employing the diffusion inversion process on the rendered images. These noisy latents encode layout priors induced by camera motion, enabling camera controllability without any finetuning or additional injection layers in the diffusion model. Let $x_0^{ref}$ and $ M^{ref}$ denote the rendered RGB and mask images from $\{V_{gen}\}^t$, where $M^{ref}$ indicates visible regions in the specified views. The noisy latents at inversion timestep $T_0$ are calculated as

\begin{figure*}[t]
    \centering
    \includegraphics[width=\linewidth]{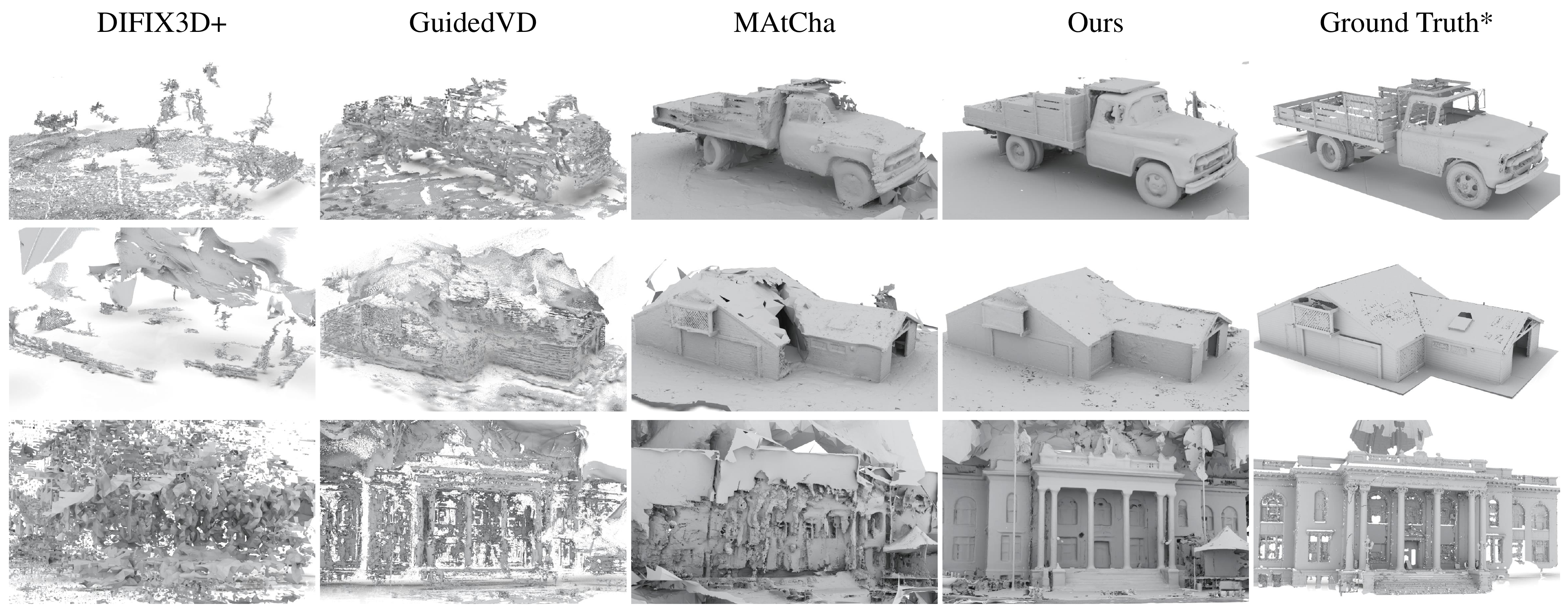}
    \vspace{-0.8cm}
    \caption{Visualization of surface reconstruction on TanksAndTemples~\cite{Knapitsch2017tanks} dataset. We obtain the GT mesh through Poisson-Disk reconstruction on the GT point clouds for reference. We produce complete and high-fidelity surfaces from only 5 input views.}
    \vspace{-0.2cm}
    \label{fig:compare-tnt}
    \Description{Visualization of surface reconstruction on TanksAndTemples dataset.}
\end{figure*}

\begin{equation}
\label{eq4}
    x_{T_0}= (1-T_0)x_0^{ref}+T_0 \epsilon, \epsilon\sim \mathcal{N}(0, I).
\end{equation}
\noindent Starting from $x_{T_0}$, we perform flow matching denoising process to obtain the clean latents $x_0$ as follows,
\begin{equation}
    x'_{t-1}=x_t-\Delta t\ v_\theta (x_t, t, c),
\end{equation}
\noindent where $v_\theta (x_t, t, c)$ denotes the estimated flow field at $x_t$. Unfortunately, such generation cannot be directly used for reconstruction, as demonstrated in Fig.~\ref{fig:ablation-denoising}, where the stochastic denoising process introduces significant content drift. To address this issue, inspired by image inpainting~\cite{lugmayr2022repaint,ju2024brushnet,lei2023rgbd2}, we utilize the rendered results in known regions as references and adjust the denoising direction toward the underlying scene geometry. Specifically, we inverse $x_0^{ref}$ to timestep $t-1$ to obtain $x_{t-1}^{ref}$ using the same process in Eq.~\ref{eq4}. We then blend $x'_{t-1}$ and $x_{t-1}^{ref}$ to obtain the adjusted noisy latents $x_{t-1}$ for the next-step denoising,
\begin{equation}
    x_{t-1}=M(t)x_{t-1}^{ref}+(1-M(t))x'_{t-1},
\end{equation}
\noindent where $M(t)$ denotes the spatial mask that controls the blending between the two noisy latents. Based on the observation that diffusion denoising establishes global semantics at early denoising stages while refines spatial details at later stages~\cite{ho2020ddpm,peng2025omnisync,wan2025wan}, we design a three-stage denoising control strategy to guide the generation:
\begin{equation}
    M(t)=\begin{cases}
        M^{ref} & T_1 < t \leq T_0 \\
        (\frac{T_1-t}{T_1-T_2})^\rho M^{ref} & T_2 < t \leq T_1 \\
        0 & 0 \leq t \leq T_2
    \end{cases}.
\end{equation}

\noindent In the early stage ($T_0\!\rightarrow\!T_1$), we enforce the denoising direction within known regions to strictly follow the rendered references, which anchors the scene structure and prevents dynamics and content drift. In the middle stage ($T_1\!\rightarrow\!T_2$), we gradually relax the constraint. In the final stage ($T_2\!\rightarrow\!0$), we unfreeze it to refine imperfect renderings and synthesize realistic local details. As shown in Fig.~\ref{fig:compare-videoconsistency}, our stage-wise denoising strategy maintains strong 3D consistency while faithfully adhering to the underlying scene geometry compared to other camera-controlled video generation methods.

\subsection{Loss Function}
\label{sec3.5}

The overall optimization objective is $\mathcal{L}=\mathcal{L}_{input}+\mathcal{L}_{gen}$.  Here $\mathcal{L}_{input}$ is used for the initial GT sparse input views $\{V_{input}\}^0$, defined as
\begin{equation}
\mathcal{L}_{input}=\mathcal{L}_{c}+\lambda_1\mathcal{L}_{reg}+\lambda_2\mathcal{L}_n,
\end{equation}
\noindent where $\mathcal{L}_c$ is the photometric loss~\cite{kerbl20233dgs}, $\mathcal{L}_{reg}$ denotes the regularization loss used in 2DGS~\cite{huang20242dgs} and MAtCha~\cite{guedon2025matcha}, and $\mathcal{L}_n=|1-N^T \hat{N}|$ is the normal prior loss between the rendered normals and monocular normals~\cite{hu2024metric3dv2}. $\mathcal{L}_{gen}$ is used for the generated views $\{V_{gen}\}^t$, defined as
\begin{equation}
\mathcal{L}_{gen}=U\odot(\mathcal{L}_{lap}+\lambda_1\mathcal{L}_{reg}+\lambda_2\mathcal{L}_n),
\end{equation}
\noindent where we replace $\mathcal{L}_c$ with a Laplacian loss~\cite{niklaus2018context} to mitigate the generated artifacts in the high-frequency details. $U$ denotes a per-pixel confidence map derived from the point cloud fusion process~\cite{wang2025vggt}.

\begin{table}[t]
  \vspace{-0.0cm}
  \centering
  \caption{\md{Numerical comparisons of sparse-view reconstruction accuracy on Replia~\cite{straub2019replica} and TanksAndTemples~\cite{Knapitsch2017tanks} datasets.}}
  \vspace{-0.3cm}
  \label{tab:replica-tnt}
  \Description{Numerical comparisons of sparse-view reconstruction accuracy on Replia and TanksAndTemples datasets.}
  \resizebox{\linewidth}{!}{
  \begin{tabular}{l|cc|ccc}
    \toprule
       Datasets & \multicolumn{2}{c|}{TNT} & \multicolumn{3}{c}{Replica}  \\
    \cmidrule{1-6}
       Methods & CD$\downarrow$ & F-Score$\uparrow$ & CD$\downarrow$ & NC$\uparrow$ & F-Score$\uparrow$  \\
    \midrule
      FSGS~\cite{zhu2024fsgs} & 0.93 & \phantom{0}3.00 & 0.12 & 67.36 & 39.46   \\
      FatesGS~\cite{huang2025fatesgs} & 1.04 & \phantom{0}1.68 & 0.74 & 52.47 & \phantom{0}3.47   \\
      PGSR~\cite{chen2024pgsr}  & 0.92 & \phantom{0}2.61 & 0.59 & 55.01 & 17.76  \\
      Sparse2DGS~\cite{wu2025sparse2dgs}  & 1.35 & \phantom{0}0.49 & 0.82 & 49.64 & \phantom{0}2.83  \\
      DIFIX3D+~\cite{wu2025difix3d+}  & 0.82 & \phantom{0}0.21 & 0.70 & 51.11 & \phantom{0}4.77  \\
      GuidedVD~\cite{zhong2025taming}  & 0.81 & \phantom{0}2.91 & 0.12 & 67.91 & 40.45  \\
      MAtCha~\cite{guedon2025matcha}  & 0.84 & \phantom{0}6.47 & 0.08 & 83.18 & 71.77  \\
      QGS~\cite{zhang2024quadratic}  & 0.93 & \phantom{0}2.72 & 0.23 & 60.73 & 28.66  \\
      Ours & \textbf{0.66} & \textbf{12.80} & \textbf{0.06} & \textbf{88.42} & \textbf{80.79}   \\
  \bottomrule
    \end{tabular}
    }
\vspace{-0.4cm}
\end{table}

\section{Experiments}

\subsection{Experimental Settings}

\begin{figure}[t]
    \centering
    \includegraphics[width=\linewidth]{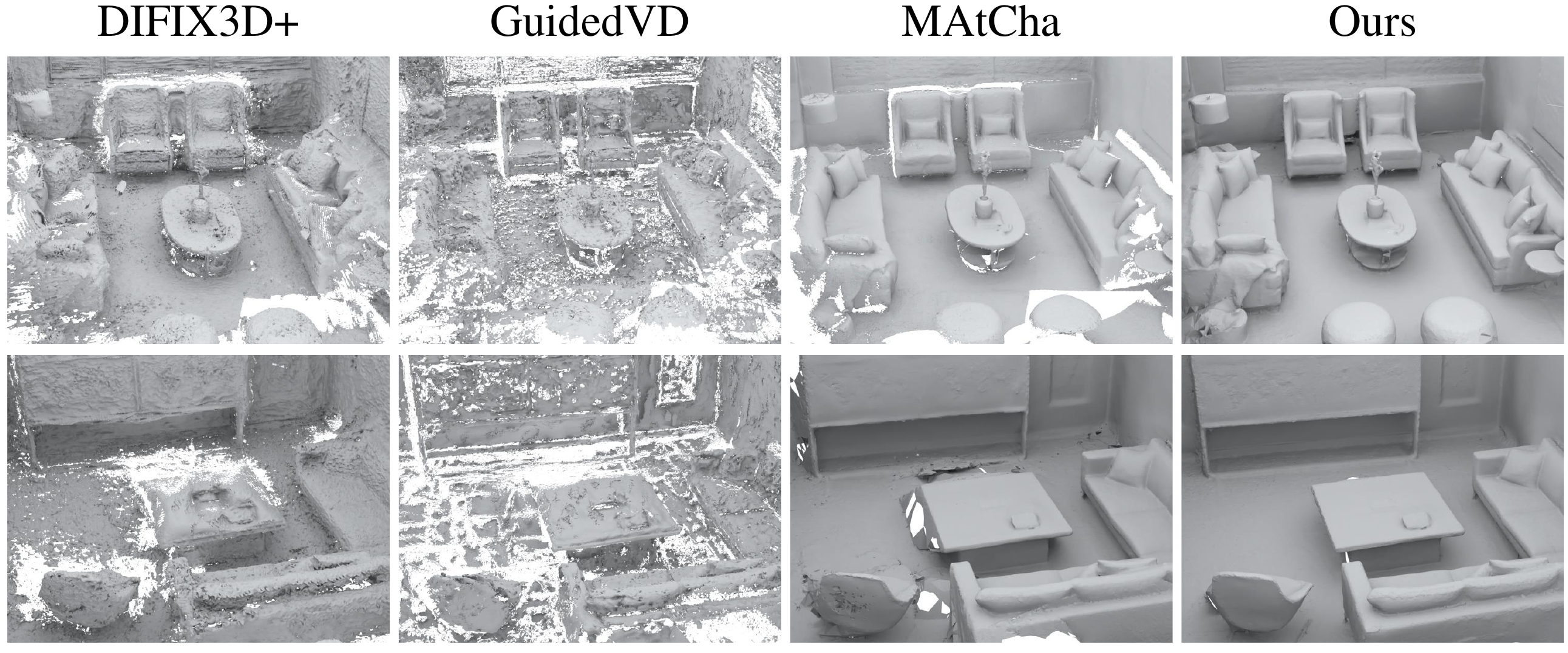}
    \vspace{-0.6cm}
    \caption{Visualization of surface reconstruction on Replica~\cite{straub2019replica} dataset with 10 input views. We are able to reconstruct delicate surfaces without holes.}
    \vspace{-0.2cm}
    \label{fig:compare-replica}
    \Description{Visualization of surface reconstruction on Replica dataset with 10 input views.}
\end{figure}

\noindent\subsubsection{Implementation Details} We adopt pretrained Wan2.1 I2V~\cite{wan2025wan} as our base video diffusion model, and use MAtCha~\cite{guedon2025matcha} as ours surface reconstruction backbone. For each camera trajectory, we sample 16 viewpoints and select 4 frames from the generated video to form $\{V_{gen}\}^t$. Our reconstruction framework is trained for a total of 15000 iterations. Starting from 7000 iteration, we perform mesh evaluation and video generation for every 4000 iterations, which cycles two rounds in total. More implementation details are provided in the supplementary materials.

\noindent\subsubsection{Datasets} We evaluate our method on three challenging datasets covering both indoor and outdoor scenarios: (1) \textbf{Tanks and Temples (TNT)}~\cite{Knapitsch2017tanks}, where we use all 6 scenes and select 5 input views per scene; (2) \textbf{Replica}~\cite{straub2019replica}, where we use all 8 scenes with 10 input views per scene; (3) \textbf{DL3DV}~\cite{ling2024dl3dv}, where we use 4 indoor and 4 outdoor scenes from its benchmark, selecting 6 input views for each. 

\begin{figure*}[t]
    \centering
    \includegraphics[width=\linewidth]{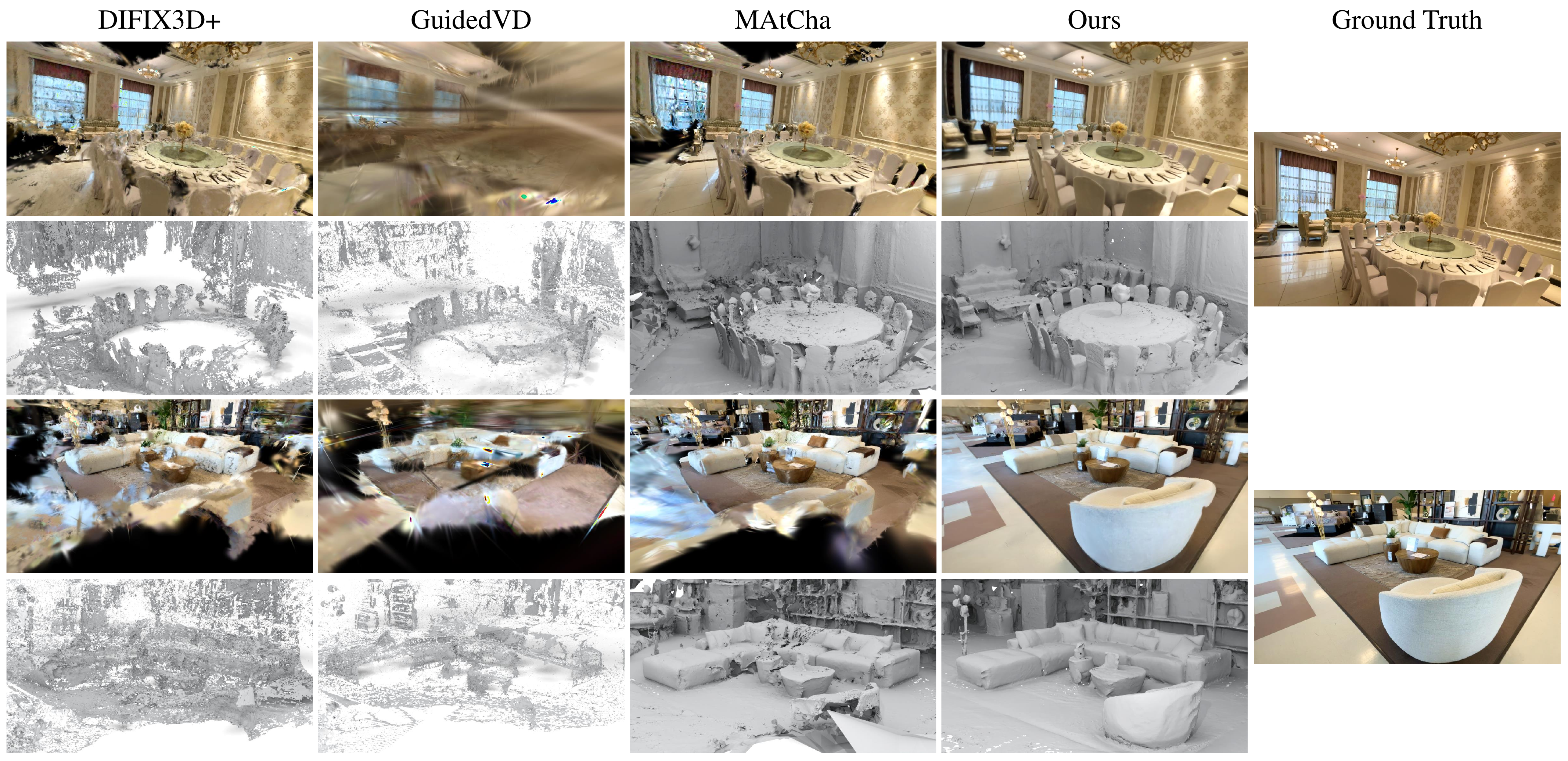}
    \vspace{-0.6cm}
    \caption{Visualization of surface reconstruction and novel view synthesis on DL3DV~\cite{ling2024dl3dv} dataset with 6 input views. Our method successfully reconstructs and renders complete scenes with high-fidelity.}
    \vspace{-0.2cm}
    \label{fig:compare-dl3dv}
    \Description{Visualization of surface reconstruction and novel view synthesis on DL3DV dataset with 6 input views.}
\end{figure*}

\noindent\subsubsection{Baselines} We compare our method with three categories of methods: (1) Dense-view reconstruction methods; (2) Sparse-view reconstruction methods; and (3) Sparse-view novel view synthesis methods with generative priors. We also evaluate the performance of video generation with other camera-controlled video diffusion methods.

\begin{figure*}[t]
    \centering
    \includegraphics[width=\linewidth]{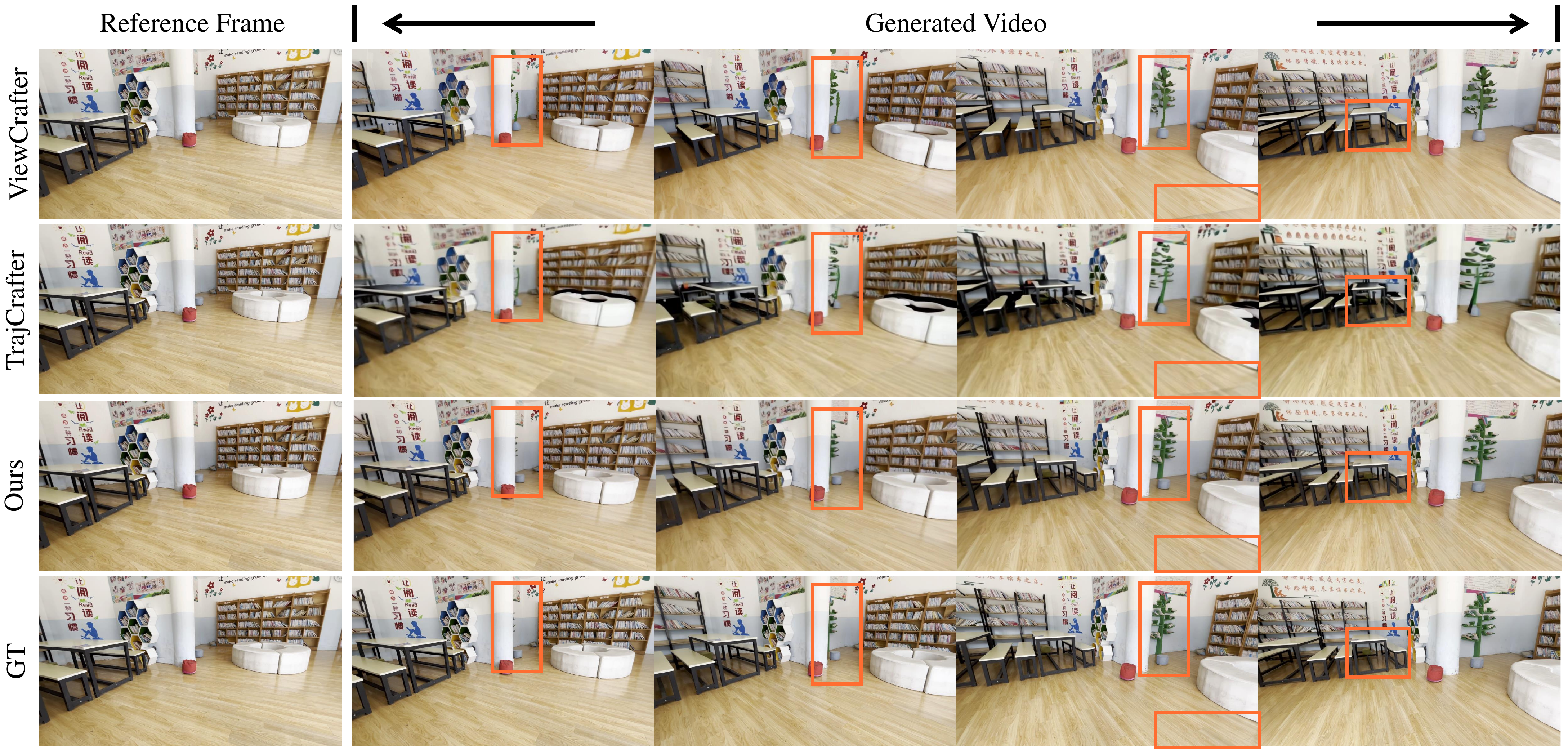}
    \vspace{-0.6cm}
    \caption{Comparison of our method with other camera-controlled video generation methods. We achieve significantly more consistent results obeying ground truth geometries.}
    \vspace{-0.2cm}
    \label{fig:compare-videoconsistency}
    \Description{Comparison of our method with other camera-controlled video generation methods.}
\end{figure*}

\subsection{Comparison Results}

\noindent\subsubsection{Surface Reconstruction} We report the quantitative results in Tab.~\ref{tab:replica-tnt} on TNT and Replica datasets, where our method achieves significantly better performance than all baselines. 
Visual comparisons in Fig.~\ref{fig:compare-tnt},~\ref{fig:compare-replica},~\ref{fig:compare-dl3dv} further demonstrate that our method can reconstruct complete surfaces with high-quality geometric details under sparse-view inputs.

\begin{table}[t]
  \vspace{-0.0cm}
  \centering
  \caption{Numerical comparisons of novel view synthesis with 6-view inputs on DL3DV~\cite{ling2024dl3dv} dataset.}
  \vspace{-0.2cm}
  \label{tab:dl3dv}
  \Description{Numerical comparisons of novel view synthesis with 6-view inputs on DL3DV dataset.}
  \resizebox{\linewidth}{!}{
  \begin{tabular}{l|ccc|ccc}
    \toprule
        & \multicolumn{3}{c|}{Indoor Scenes} & \multicolumn{3}{c}{Outdoor Scenes} \\
    \cmidrule{1-7}
       Methods & PSNR$\uparrow$ & SSIM$\uparrow$ & LPIPS$\downarrow$ & PSNR$\uparrow$ & SSIM$\uparrow$ & LPIPS$\downarrow$ \\
    \midrule
      FSGS & 16.30 & 0.622 & 0.411 & 15.56 & 0.507 & 0.435  \\
      FatesGS & 14.07 & 0.505 & 0.454 & 13.81 & 0.412 & 0.444  \\
      PGSR & 16.31 & 0.616 & 0.386 & 15.61 & 0.544 & 0.426  \\
      Sparse2DGS & 12.96 & 0.276 & 0.535 & 13.12 & 0.260 & 0.593  \\
      DIFIX3D+ & 16.36 & 0.549 & 0.391 & 14.66 & 0.422 & 0.419  \\
      GuidedVD & 17.87 & 0.649 & 0.387 & 17.02 & 0.516 & 0.43  \\
      MAtCha & 16.83 & 0.598 & 0.326 & 15.23 & 0.453 & 0.384  \\
      QGS & 16.73 & 0.599 & 0.359 & 15.35 & 0.526 & 0.387  \\
      Ours & \textbf{19.78} & \textbf{0.699} & \textbf{0.292} & \textbf{17.49} &  \textbf{0.561} & \textbf{0.376}  \\
  \bottomrule
    \end{tabular}
    }
\vspace{-0.2cm}
\end{table}

\begin{table}[t]
  \vspace{-0.0cm}
  \centering
  \caption{\md{Quality evaluation of generated videos between baselines, our method and our ablations. We report both rendering metrics on images and generation metrics on videos.}}
  \vspace{-0.2cm}
  \label{tab:video-generation}
  \Description{Quality evaluation of generated videos between baselines, our method and our ablations.}
  \resizebox{\linewidth}{!}{
  \begin{tabular}{l|ccccc}
    \toprule
       Methods & PSNR$\uparrow$ & SSIM$\uparrow$ & LPIPS$\downarrow$ & FID$\downarrow$ & FVD$\downarrow$ \\
    \midrule
      CamTrol~\cite{hou2025camtrol} & 14.53 & 0.639 & 0.462 & 110.42 & 625.99  \\
      ViewCrafter~\cite{yu2024viewcrafter} & 23.85 & 0.810 & 0.297 & 60.78 & 178.93  \\
      TrajectoryCrafter~\cite{mark2025trajectorycrafter} & 21.42 & 0.781 & 0.323 & 74.04 & 164.28  \\
      DIFIX3D+~\cite{wu2025difix3d+} & 20.14 & 0.762 & 0.301 & 66.11 & 220.08  \\
      Ours & \textbf{25.80} & \textbf{0.847} & \textbf{0.238} & \textbf{56.42} & \textbf{114.42} \\
    \cmidrule{1-6}
      Ours (SVD) & 21.80 & 0.740 & 0.324 & 59.96 & 169.52 \\
      Ours (Hunyuan 1.0) & 24.40 & 0.808 & 0.292 & 62.64 & 119.54 \\
    \cmidrule{1-6}
      Ours (w/o Guiding) & 15.67 & 0.691 & 0.382 & 89.24 & 261.97 \\
      Ours (w/o Mask) & 21.02 & 0.798 & 0.266 & 84.48 & 231.20    \\
      Ours (Only Stage1) & 24.30 & 0.835 & 0.265 & 75.40 & 241.04  \\
  \bottomrule
    \end{tabular}
    }
\vspace{-0.3cm}
\end{table}

\noindent\subsubsection{Novel View Synthesis} We further evaluate novel view synthesis on DL3DV dataset, as reported in Tab.~\ref{tab:dl3dv}, where our method consistently achieves the best results across both indoor and outdoor scenes. Visual comparisons in Fig.~\ref{fig:compare-dl3dv} show that our method can produce high quality renderings in regions that are sparsely or not covered by the input views.

\noindent\subsubsection{Video Generation} We further evaluate the video generation performance conditioned on rendering results and camera trajectories by comparing the generated videos with ground truth sequences, as reported in Tab.~\ref{tab:video-generation}. Compared with advanced camera-controlled video generation methods~\cite{yu2024viewcrafter,mark2025trajectorycrafter,hou2025camtrol}, our method achieves superior consistency between rendered and real results (PSNR, SSIM, LPIPS), as well as greater generative diversity (FID, FVD). As visualized in Fig.~\ref{fig:compare-videoconsistency}, baseline methods either fail to fill missing regions or produce hallucinated content, whereas our method, explicitly guided by rendering-based denoising directions, generates high-fidelity results that faithfully adhere closely to the underlying geometry.

\subsection{Application on Single-View Generation}
We further present an application of our method on single-view generation. Given a single input view, we first estimate monocular metric depth~\cite{hu2024metric3dv2} and backproject it into a 3D point cloud to initialize Gaussian primitives. During Gaussian training, we iteratively construct orbiting camera trajectories that progressively expand both the Gaussian primitives $\{G_i\}$ and the training views $\{V_{input}\}^t$ until most of the scene is covered. Fig.~\ref{fig:teaser} shows single-view reconstruction results of an object from DTU dataset and an indoor scene from AIGC, where our method successfully recovers large regions that are invisible in the input image. This highlights the strong generalization capability of our approach.

\begin{figure}[t]
    \centering
    \includegraphics[width=\linewidth]{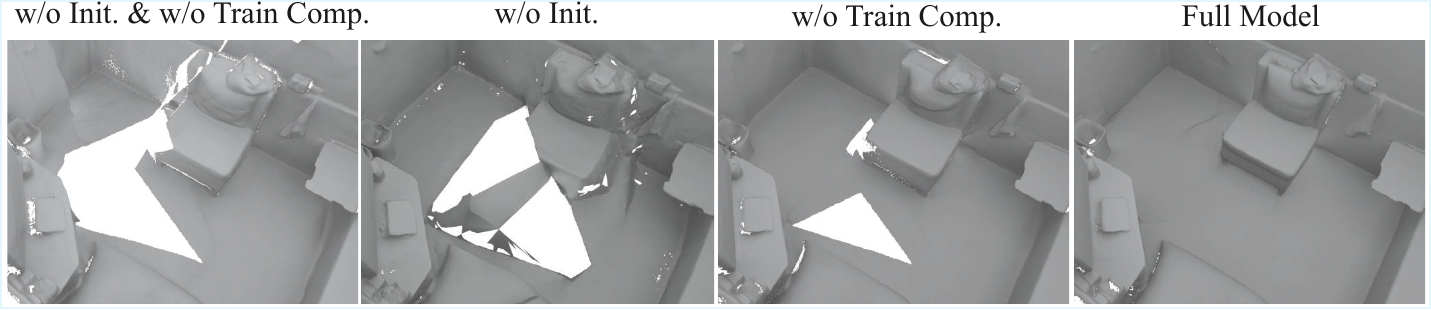}
    \vspace{-0.6cm}
    \caption{\md{Ablation study on our effectiveness of the initialization completion and training completion modules.}}
    \vspace{-0.2cm}
    \label{fig:ablation-pointcloud-train}
    \Description{Ablation study on our effectiveness of the initialization completion and training completion modules.}
\end{figure}

\subsection{Ablation Studies}
In this section, we present ablation studies on the effectiveness of each one of our modules. Additional ablations, comparisons and analysis are available in the supplementary materials.

\noindent\subsubsection{Effectiveness of completion modules of the framework} We first validate the effectiveness of our completion modules, as reported in Tab.~\ref{tab:init-train-module}. \md{``Initialization completion'' means using video priors to generate novel views for completing the initial point cloud. ``Training completion'' means iteratively using video priors to expand the set for training Gaussian Splatting.} Visual results in Fig.~\ref{fig:ablation-pointcloud-train} show that without completion (\textbf{w/o Init. \& w/o Train Comp.}), the reconstructed scenes contain many holes. \md{Introducing initialization completion while excluding training completion (\textbf{w/o Train Comp.}) alleviates this issue but still leaves noticeable holes, since it is difficult to expand the trajectories to cover the full scene merely from sparse input views. Similarly, when training completion is applied without expanding the initial point cloud (\textbf{w/o Init. Comp.}), the reconstructed result is also poor due to the lack of sufficient geometric priors at the very beginning. When combined with both initialization and training completion (\textbf{Full Model}), our framework recovers complete and high-quality surfaces. }

\noindent\subsubsection{Effectiveness of the stage-wise denoising} 
\begin{figure}[t]
    \centering
    \includegraphics[width=\linewidth]{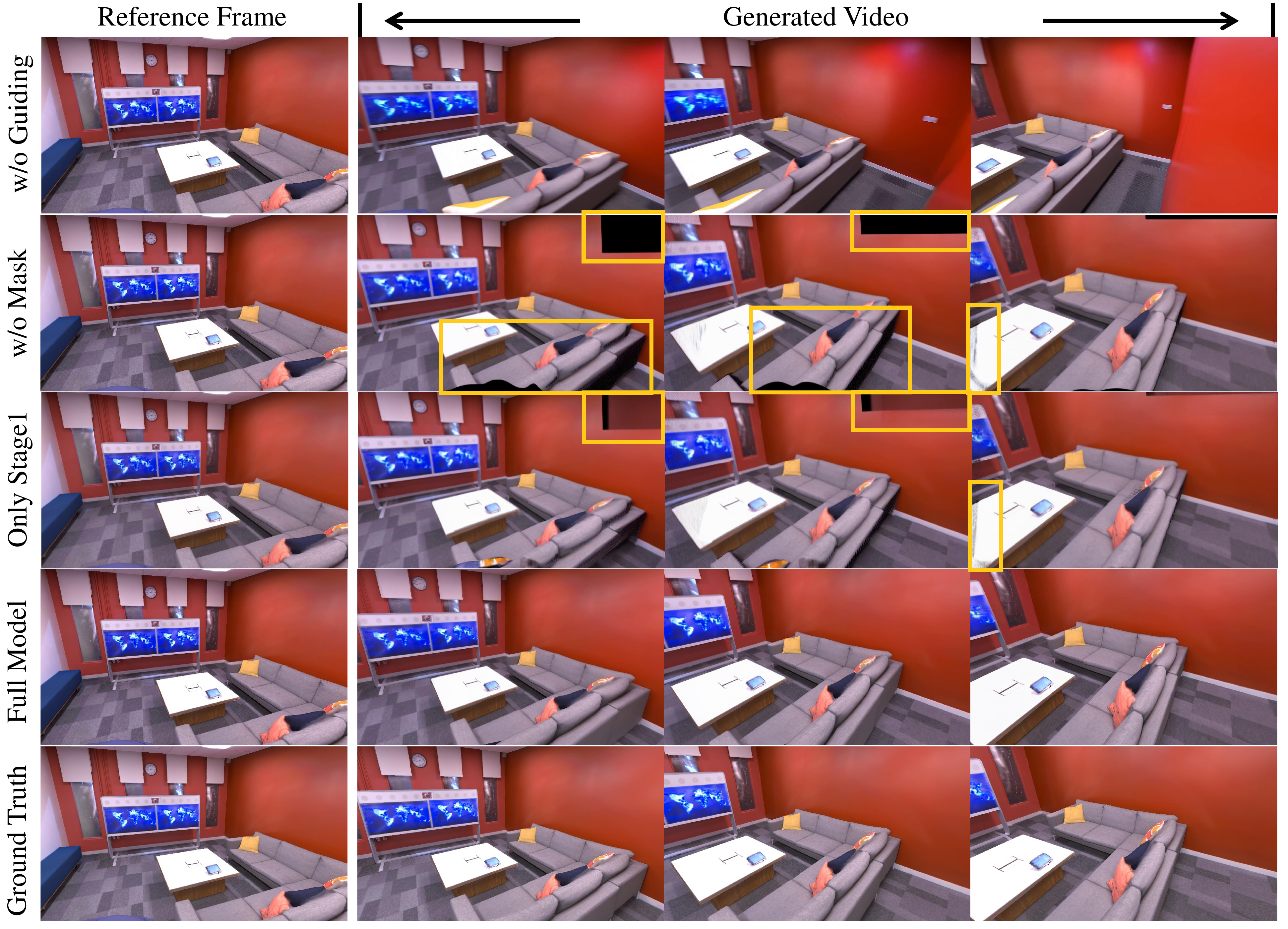}
    \vspace{-0.6cm}
    \caption{Ablation study on our stage-wise denoising strategy.}
    \vspace{-0.2cm}
    \label{fig:ablation-denoising}
    \Description{Ablation study on our stage-wise denoising strategy.}
\end{figure}

We report ablation results on our denoising strategy in Tab.~\ref{tab:video-generation} and Fig.~\ref{fig:ablation-denoising} both quantitatively and qualitatively. Without geometry guiding (\textbf{w/o Guiding}), that is, denoising the inverse noisy latents without any control, the generated videos appear visually plausible but exhibit noticeable content misalignment and 3D inconsistency. If the denoising direction within the masked region keeps following the rendering results during the entire denoising process (\textbf{Only Stage 1}), the outputs fail to repair rendering artifacts, and fail to fill missing regions. The observations are similar when mask images are omitted (\textbf{w/o Mask}), that is, using the full rendered RGB images as guiding reference. With our stage-wise denoising strategy, we obtain high-fidelity videos that are consistent with the real sequences.

\noindent\subsubsection{Effectiveness of the video diffusion backbones}
We further validate the effect of different video backbones in our framework. Specifically, we replace Wan 2.1 with earlier diffusion backbones, including HunyuanVideo 1.0~\cite{kong2024hunyuanvideo} and SVD~\cite{blattmann2023stable}. As reported in Tab.~\ref{tab:video-generation}, our method maintains strong performance and notably outperforms existing methods even when using earlier architectures (e.g., SVD). This demonstrates that our geometry-guided generation strategy is highly robust and effectively unleashes the capacity of various video foundation models without requiring model-specific finetuning.

\noindent\subsubsection{Exposure consistency}
\begin{figure}[t]
    \centering
    \includegraphics[width=\linewidth]{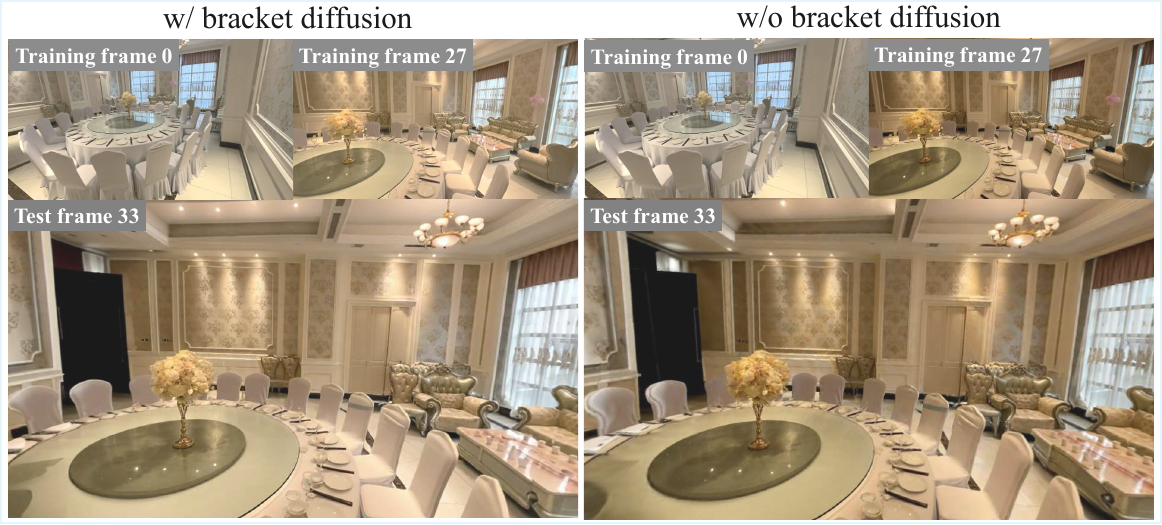}
    \vspace{-0.6cm}
    \caption{\md{Ablation study on exposure consistency preprocessing using BracketDiffusion.}}
    \vspace{-0.2cm}
    \label{fig:ablation-exposure-consistency}
    \Description{Ablation study on exposure consistency preprocessing using BracketDiffusion.}
\end{figure}

\md{Real-world sparse views often suffer from varying exposures, which can disrupt photometric consistency. To address this, we conducted an additional experiment using BracketDiffusion~\cite{bemana2025bracket} to convert the input images into exposure-consistent images for training. Numerical results reported in Tab.~\ref{tab:bracket-diffusion} demonstrate the effectiveness of our approach. As further illustrated in Fig.~\ref{fig:ablation-exposure-consistency}, even when the training views exhibit significant exposure inconsistencies, our preprocessing enables the model to produce more harmonious lighting in novel views. This indicates that enhancing exposure consistency provides a more robust photometric constraint for our optimization process.}

\begin{table}[t]
  \centering
  \caption{\md{Effectiveness of exposure consistency preprocessing using BracketDiffusion (BD).}}
  \vspace{-0.2cm}
  \label{tab:bracket-diffusion}
  \Description{Effectiveness of exposure consistency preprocessing using BracketDiffusion (BD).}
  \begin{tabular}{l c c c}
    \toprule
    Methods & PSNR$\uparrow$ & SSIM$\uparrow$ &  LPIPS$\downarrow$ \\
    \midrule
    w/o BracketDiffusion & 19.748 & \textbf{0.7541} & 0.2623 \\
    w/ BracketDiffusion & \textbf{19.872} & 0.7466 & \textbf{0.2548} \\
    \bottomrule
  \end{tabular}
  \vspace{-0.2cm}
\end{table}

\begin{table}[t]
  \vspace{-0.0cm}
  \centering
  \caption{\md{Ablation study on the initialization and training completion modules of our framework on Replica dataset.}}
  \vspace{-0.2cm}
  \label{tab:init-train-module}
  \Description{Ablation study on the initialization and training completion modules of our framework on Replica dataset.}
  \begin{tabular}{c c c c}
    \toprule
    Init. Comp. & Train Comp. & CD$\downarrow$ & F-Score$\uparrow$ \\
    \midrule
    \xmark & \xmark & 0.0784 & 72.100 \\
    \cmark & \xmark & 0.0711 & 74.490 \\
    \xmark & \cmark & 0.0674 & 77.923 \\
    \cmark & \cmark & \textbf{0.0631} & \textbf{80.797} \\
    \bottomrule
  \end{tabular}
  \vspace{-0.1cm}
\end{table}

\section{Conclusion}
In this work, we introduced \md{VidSplat}, a generative sparse-view reconstruction framework that integrates geometry-guided video diffusion priors with Gaussian Splatting to recover complete and high-fidelity 3D scenes from limited inputs. \md{VidSplat} addresses two key challenges in generative sparse-view reconstruction. First, we improve the 3D consistency of video generation through a training-free, stage-wise denoising strategy. Second, we develop an iterative optimization framework that progressively expands scene coverage for complete reconstruction. Extensive experiments on diverse real-world benchmarks show that \md{VidSplat} significantly outperforms existing sparse-view reconstruction and generative novel view synthesis methods in both geometry accuracy and rendering fidelity. Moreover, \md{VidSplat} exhibits strong generalization ability, enabling promising applications such as single-image reconstruction.

\begin{acks}
This work was partially supported by Deep Earth Probe and Mineral Resources Exploration -- National Science and Technology Major Project (2024ZD1003405), and the National Natural Science Foundation of China (62272263), and in part by Kuaishou.
\end{acks}

\bibliographystyle{ACM-Reference-Format}
\bibliography{sample-base}


\end{document}